\documentclass[letterpaper,10pt,conference]{ieeeconf}
\IEEEoverridecommandlockouts
\usepackage{amsmath,amssymb,booktabs,graphicx,xcolor,url,multirow}
\usepackage{algorithm,algpseudocode}
\usepackage[hidelinks]{hyperref}
\usepackage{microtype}
\usepackage{stfloats}
\usepackage{cleveref}
\crefname{figure}{Fig.}{Figs.}
\Crefname{figure}{Figure}{Figures}
\crefname{equation}{Eq.}{Eqs.}
\Crefname{equation}{Equation}{Equations}
\crefname{table}{Table}{Tables}
\Crefname{table}{Table}{Tables}
\newcommand{\vect}[1]{\boldsymbol{#1}}
\newcommand{\method}{DeCOD}

\title{\LARGE\bf  Degeneracy-Orthogonal Geometric Constraints for LiDAR SLAM}
\author{Minseo Kim$^{*}$, Yina Kim$^{*}$, Jinhwa Hwang, and Alex Junho Lee$^{\dagger}$%
\thanks{$^{*}$Minseo Kim and Yina Kim contributed equally to this work.}%
\thanks{All authors are with the Department of Mechanical Systems Engineering, Sookmyung Women's University, 100 Cheongpa-ro 47-gil, Yongsan-gu, Seoul 04310, Republic of Korea.}%
\thanks{\raggedright $^{\dagger}$Corresponding author: Alex Junho Lee (alexjunholee@gmail.com).}}
\begin{document}
\maketitle
\enlargethispage{\baselineskip}
\thispagestyle{empty}\pagestyle{empty}
\begin{abstract}
Autonomous robot navigation relies on simultaneous localization and mapping (SLAM) to estimate motion and maintain an accurate pose within an environment.
However, in axially uniform corridors such as long tunnels and pipelines, LiDAR odometry is fundamentally limited by unconstrained drift along the feature-weak travel direction. This structural degeneracy cannot be resolved by local scan matching alone.
To address this challenge, we propose the Degeneracy-orthogonal Contour Offset Descriptor (\method{}), a structure-orthogonal geometric descriptor for cross-sectional landmarks.
Cross-sectional boundaries, such as pipe joints and structural rings, provide metric constraints along this degenerate axis, but distinguishing individual landmarks requires capturing subtle surface variations across nearly identical profiles.
The descriptor parameterizes signed normal deviation from estimated boundary contours, and matching explicitly resolves heading ambiguity and compensates for first-order contour errors.
Matched landmarks yield geometric factors that enforce agreement in cross-section position and corridor axis alignment during pose-graph optimization, correcting longitudinal drift while leaving rotation about the common axis unconstrained.
On a public benchmark and in field experiments, \method{} achieves robust landmark retrieval over standard 3D descriptors and successfully stabilizes trajectories across different odometry frontends, reliably constraining longitudinal drift under geometric degeneracy.
\end{abstract}

\section{Introduction}

Subterranean environments, such as sewer pipes, subway tunnels, and mines, require autonomous robotic inspection due to hazardous and confined conditions.
In these inspection tasks, maintaining an accurate estimate of a robot's position is essential to pinpoint structural defects, such as cracks or leaks, in GPS-denied passages.
While LiDAR-based simultaneous localization and mapping (SLAM) estimates motion by registering successive observations~\cite{zhang2014loam,shan2020liosam}, geometric degeneracy can lead to erroneous motion estimates.
In long corridors such as tunnels, surfaces are repeated along their longitudinal axis.
Translation along this degenerate direction is weakly constrained, allowing significant position errors to accumulate despite locally consistent surface alignment~\cite{zhang2016degeneracy,tuna2022xicp}.

Degeneracy-aware estimators mitigate divergence by restricting unreliable updates~\cite{zhang2016degeneracy,tuna2022xicp}, but recovering from accumulated drift fundamentally requires additional measurements.
While external infrastructure (e.g., ultra-wideband anchors) can provide position constraints~\cite{zhen2019tunnel}, fully autonomous operation requires re-identifying previously visited structures upon revisits.
However, standard place recognition methods~\cite{kim2021scancontextpp,yuan2023std} struggle in these environments due to severe structural ambiguity caused by repetitive geometric layouts.

\begin{figure}[t]
	\centering
	\includegraphics[width=0.95\linewidth]{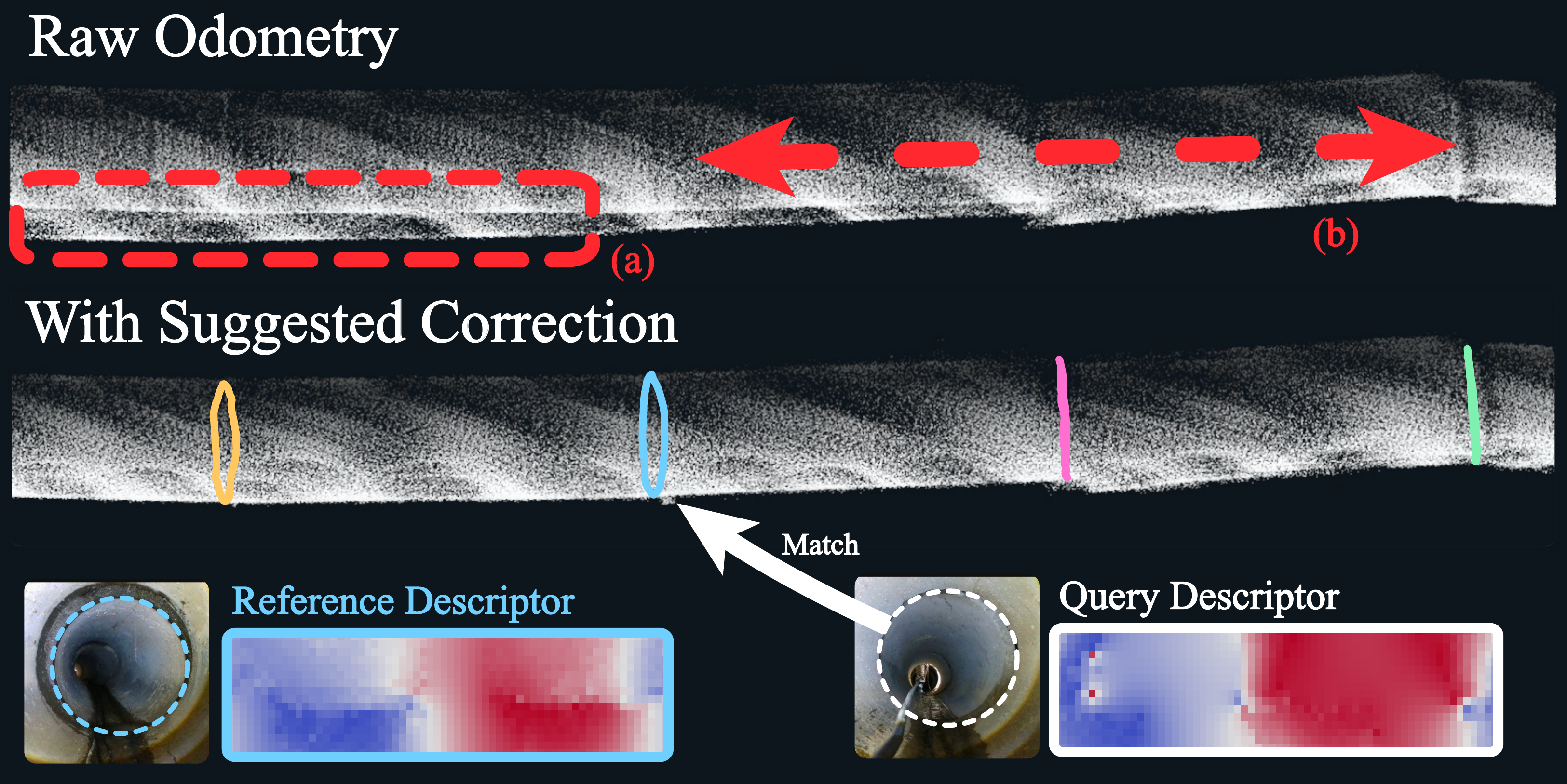}
	\caption{Overview of \method{} under geometric degeneracy. In raw LiDAR odometry (top), uniform walls cause severe longitudinal drift (a) along the degenerate corridor axis (b). To constrain this drift, \method{} extracts cross-sectional boundaries, shown as colored rings (middle), and unwraps them into 2D descriptors (bottom) to supply metric weak-direction constraints.}
	\label{fig:overview}
\vspace{-6mm}
\end{figure}

For localization under geometric degeneracy, an association must provide constraints sensitive to motion along the weak direction.
While axially uniform walls reliably constrain lateral motion, their lack of longitudinal variation leaves the feature-weak travel axis unconstrained.
Meanwhile, cross-sectional boundaries (e.g., pipe joints, segment seams, structural rings), modeled as transverse planes, provide displacement constraints along this degenerate axis (\Cref{fig:overview}).
Directly tracking their fine surface variations within real-time odometry is computationally demanding and prone to local minima given structural repetition.
This motivates a decoupled formulation that re-identifies cross-sectional landmarks using distinctive surface descriptors and uses the resulting geometric constraints to correct the accumulated longitudinal drift.

In this paper, we propose the Degeneracy-orthogonal Contour Offset Descriptor (\method{}) to support state estimation in geometrically degenerate environments.
\method{} identifies landmark boundary contours orthogonal to the degenerate direction and encodes signed normal deviations of the surrounding surface.
The proposed method handles contour estimation errors and sensor artifacts while remaining consistent under reverse traversals.
The main contributions of this paper are summarized below.
\begin{itemize}\setlength{\itemsep}{0pt}\setlength{\parskip}{0pt}\setlength{\parsep}{0pt}
    \item A degeneracy-orthogonal geometry descriptor that encodes normalized structural deviations around cross-sectional boundary surfaces.
    \item A robust association formulation that handles contour estimation errors, partial occlusion, and reversed traversals, with geometric verification to improve precision.
    \item Experimental validation on a public benchmark and in field pipe experiments, demonstrating reliable landmark association and reduced drift under severe geometric degeneracy.
\end{itemize}

\section{Related Work}
\label{sec:related}

\subsection{Place Recognition under Degeneracy}
LiDAR place recognition establishes correspondences across observations by summarizing geometry into viewpoint-invariant descriptors.
Scan Context~\cite{kim2018scancontext} bins egocentric surface heights in a polar grid and resolves heading changes through circular shifts.
Subsequent structural methods aggregate geometric primitives, such as triangles formed from 3D keypoints in STD~\cite{yuan2023std} or elevation-sliced bird's-eye-view contours in Contour Context~\cite{jiang2023contour}.

These formulations rely on salient geometric variations throughout the surrounding environment.
In corridors and tunnel-like passages, the cross-sectional profile remains uniform along the travel axis over long distances.
Consequently, global descriptors produce near-identical signatures across distinct locations, inducing severe perceptual aliasing and false loop closures~\cite{chen2024geode,kim2025admission}.
Under 1D geometric degeneracy, resolving longitudinal position effectively reduces place recognition to associating discrete structural landmarks along the passage.
Where individual transverse structures, such as seams, boundary edges, and geometric transitions, can be extracted, landmark association provides localized spatial references that resolve longitudinal position directly from local geometry.

\subsection{Degeneracy in LiDAR Odometry}
\begin{figure}[!t]
	\centering
	\includegraphics[width=0.95\linewidth]{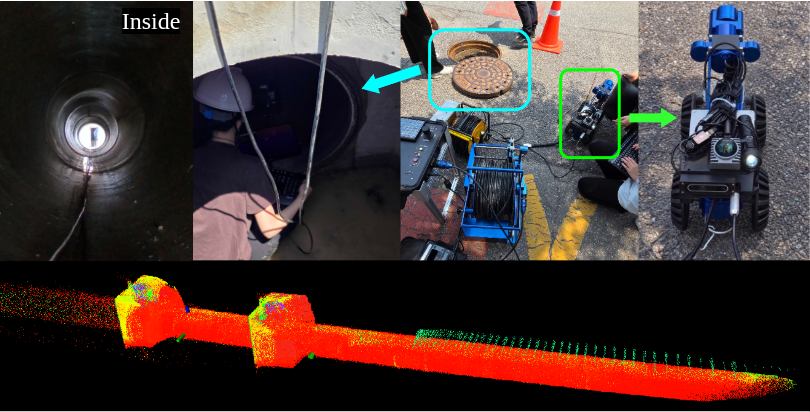}
	\caption{Field robotic inspection in underground pipes and odometry failure. Top: field setup showing (from left to right) the pipe interior, underground manhole chamber, surface setup with the manhole cover (cyan box), and the wheeled inspection robot (green box). Bottom: reconstructed point cloud exhibiting severe map compression and trajectory distortion caused by unconstrained longitudinal drift in raw LiDAR odometry.}
	\label{fig:motivation}
\vspace{-6mm}
\end{figure}

LiDAR and LiDAR-inertial odometry estimate ego-motion by registering successive surface measurements~\cite{zhang2014loam,xu2021fastlio2}.
In straight corridors and tunnels, wall surfaces run parallel to the direction of motion, so geometric registration along the travel axis becomes unconstrained~\cite{gelfand2003sampling,zhang2016degeneracy}.
Additional surface information can help constrain weakly constrained directions, such as return intensity in COIN-LIO~\cite{pfreundschuh2024coin} or subtle bump geometry in BIEVR-LIO~\cite{pfreundschuh2026bievr}.
In extended and nominally uniform passages, however, such surface variations can remain sparse or repetitive, leaving residual uncertainty along the travel axis.
Without sufficient longitudinal constraints, pose estimation can accumulate severe longitudinal drift, leading to map distortion~\cite{zhang2016degeneracy,tuna2022xicp,chen2024geode,hatleskog2024probabilistic} (\Cref{fig:motivation}).

To prevent the estimator from diverging, degeneracy-aware methods analyze the linearized system Hessian to suppress or constrain unreliable state updates~\cite{zhang2016degeneracy,tuna2022xicp,hatleskog2024probabilistic}, or construct partially constrained loop-closure factors in pose graphs~\cite{hinduja2019factors}.
External beacons such as ultra-wideband anchors~\cite{zhen2019tunnel} can supply absolute position constraints, but require pre-installed infrastructure that is impractical in extended or unexplored tunnels.
While restricting updates to observable directions maintains estimator stability, it does not provide additional geometric information along unobservable directions~\cite{zhang2016degeneracy,hinduja2019factors}.
When longitudinal constraints remain insufficient, odometry alone cannot reliably correct accumulated drift.
Bounding this error in autonomous exploration requires loop closures that supply metric constraints specifically along the travel axis upon a revisit.

Underground pipes and conduits present a representative setting for such degeneracy, where prior work leverages domain-specific models or structural elements.
Cylindrical models~\cite{zhang2021cylindrical} enforce radial surface constraints, but these constraints alone do not determine axial position.
Other approaches localize against surveyed facility maps in highway tunnels~\cite{kim2020facility} or pipe joints in sewers~\cite{edwards2023sewer}.
However, the joint-counting approach in~\cite{edwards2023sewer} assumes uniform pipe lengths and uses known manhole locations to correct errors from missed or false joint detections.
Recognizing the physical cross-section of individual structural landmarks provides metric position constraints directly from local geometry, avoiding heuristic counting and external infrastructure.

\begin{figure*}[!t]
	\centering
	\includegraphics[width=\textwidth]{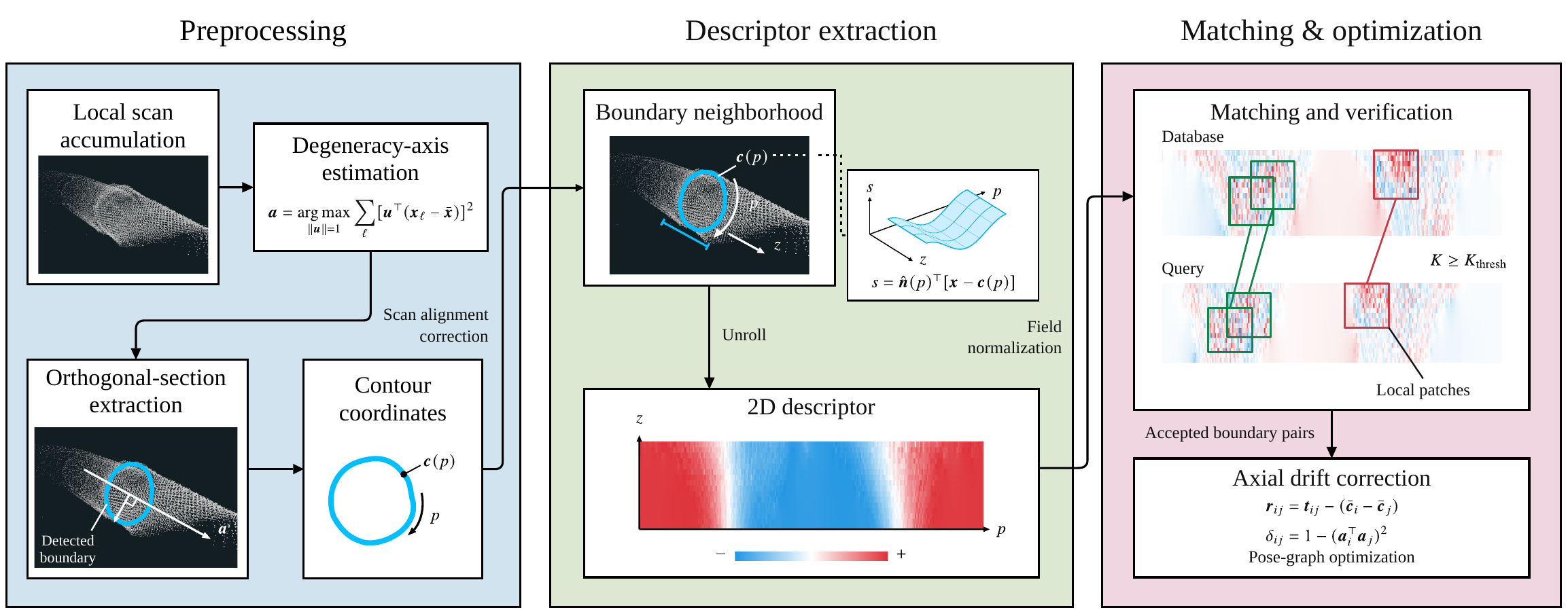}
	\caption{Overview of the \method{} pipeline in degenerate corridors. Preprocessing (left) estimates the corridor axis $\vect a$, detects structural boundaries, and fits reference contours $\vect c(p)$. In the axis estimation, $\vect x_\ell$ are input points, $\bar{\vect x}$ their mean, and $\vect u$ a unit direction. Descriptor extraction (middle) unrolls each boundary neighborhood into a 2D descriptor on the $(z,p)$ grid, where each observed cell stores the offset $s$. The matching and optimization stage (right) compensates for contour estimation errors, matches local patches, and applies geometric verification. It selects a unique Top-1 boundary candidate and accepts it only if the verification criterion is satisfied. Accepted boundary pairs supply position and axis alignment constraints for pose-graph optimization.}
	\label{fig:pipeline}
\end{figure*}

\subsection{Degeneracy-Aware Landmarks and Matching}
Representing and matching closed contour geometry has well-established foundations in shape analysis and biometrics.
Shape Context~\cite{belongie2002shapecontext} models relative point distributions for contour correspondence, while Daugman~\cite{daugman2004iris} unwrapped annular iris patterns into normalized coordinates to resolve rotational alignment via cyclic shifts and mask occlusions.
LiDAR Iris~\cite{wang2020lidariris} adapted this unwrapping principle to point clouds, generating binary signatures from surface height distributions for place retrieval.
These formulations establish that coordinate unwrapping simplifies rotational alignment.

However, directly applying existing annular representations to physical landmarks in pipes and tunnels exposes practical limitations.
Coarse binary codes and spatial histograms discard the subtle surface variations needed to distinguish nominally identical structures.
In addition, robotic inspection introduces severe physical perturbations, including partial visibility, contour estimation errors, and reversed traversal directions.
Reliable landmark association under 1D degeneracy requires resolving these challenges through continuous surface description and error-tolerant matching (\Cref{sec:method}).

\section{Drift Correction under Degeneracy}
\label{sec:method}

In environments such as tunnels and pipelines, LiDAR odometry experiences geometric degeneracy.
While surrounding wall surfaces constrain lateral motion and orientation depending on the cross-sectional geometry, their axial uniformity leaves translation along the corridor axis unconstrained.
Along this geometrically degenerate axis, local scan matching cannot measure longitudinal displacement, causing pose drift to accumulate.

To bound this longitudinal drift, \method{} associates discrete cross-sectional boundaries from axial surface profile variations, which are geometrically parameterized as transverse landmark planes defined by the detected boundary origins and the corridor axis.

To detect, describe, and associate these landmarks for drift correction, \method{} proceeds in three stages, summarized in \Cref{fig:pipeline}.
First, \Cref{sec:descriptor} extracts structural boundaries and encodes continuous signed normal offsets into degeneracy-orthogonal descriptors.
Second, \Cref{sec:invariant_matching} compensates for contour estimation errors and matches descriptors across different viewpoints.
Third, \Cref{sec:verification} verifies candidate associations and supplies position and axis alignment constraints for pose-graph optimization.

\subsection{Descriptor Extraction}
\label{sec:descriptor}
\label{sec:detection}

We construct submaps by accumulating consecutive LiDAR scans over short intervals using local odometry.
For each accumulated point cloud $\mathcal{P} \subset \mathbb{R}^3$, we estimate the longitudinal corridor axis $\vect a \in \mathbb{S}^2$ and fit a reference contour $\vect c(p)$ to observed projections onto the transverse plane orthogonal to $\vect a$~\cite{kaess2015planes}.
Structural boundaries introduce local geometric variations along the otherwise uniform corridor.

The number of degenerate degrees of freedom (DoF) depends on the geometry of the passage.
In semi-circular or horseshoe tunnels, only the translational DoF along the longitudinal axis $\vect a$ is degenerate.
In contrast, circular cylindrical pipelines exhibit two degenerate DoFs corresponding to longitudinal translation along $\vect a$ and axial rotation about $\vect a$.
To parameterize the contour without a rotational search, we define an orthonormal basis $(\vect e_1, \vect e_2)$ in the transverse plane.
Let $\vect g$ denote the gravity direction obtained from inertial measurement unit (IMU) measurements or an axis of the sensor coordinate frame.
We construct the basis as follows:
\begin{equation}
\vect e_1 = \frac{(I - \vect a \vect a^\top)\vect g}{\|(I - \vect a \vect a^\top)\vect g\|}, \qquad \vect e_2 = \vect a \times \vect e_1.
\label{eq:contour_frame}
\end{equation}

The basis fixes the circumferential origin at $p = 0$, providing a consistent contour parameterization across revisits.

To detect candidate landmarks from axial surface variations, we apply a Difference of Gaussians (DoG) filter and spatial non-maximum suppression to select distinct landmarks and their origins $\vect o$.

Inspired by annular unwrapping in iris recognition~\cite{daugman2004iris} and shape analysis~\cite{belongie2002shapecontext}, we unroll the local point cloud $\mathcal{P}$ around the fitted contour into a 2D offset chart.
For each point $\vect x \in \mathcal{P}$, the axial coordinate is $z = \vect a^\top(\vect x - \vect o)$, where $\vect o$ denotes the landmark origin, while $p \in [0, 1)$ indexes normalized circumferential arclength along $\vect c(p)$.
The signed offset
\[
s = \hat{\vect n}(p)^\top [\vect x - \vect c(p)]
\]
measures inward and outward surface variations along the outward unit normal $\hat{\vect n}(p)$.
The transverse normal components are $n_1(p) = \hat{\vect n}(p)^\top \vect e_1$ and $n_2(p) = \hat{\vect n}(p)^\top \vect e_2$.

We define the descriptor matrix $\vect D \in \mathbb{R}^{N_z \times N_p}$ by discretizing the $(z, p)$ chart into $N_z$ axial bins and $N_p$ circumferential bins.
Each observed cell stores the mean signed normal offset of its contributing points.
Recording these offsets around each landmark preserves fine surface variations that distinguish boundaries with similar cross-sectional profiles.

\subsection{Invariant Matching}
\label{sec:invariant_matching}

Once descriptors are extracted, the key challenge is associating query observations $i$ with previously visited landmarks $j$.
In degenerate corridors where accumulated longitudinal drift precludes spatial search windows, candidate retrieval relies on descriptor similarity.
However, directly comparing raw descriptor matrices $\vect D_i$ and $\vect D_j$ is unreliable because of opposite travel directions, partial surface coverage, and minor boundary misalignment.
Therefore, we define a distance measure that compares surface geometry over mutually observed regions.

First, when a robot revisits a passage in the opposite direction, its heading along the corridor axis reverses by $180^\circ$.
The reversal negates the axial coordinate $z$ and reverses the circumferential traversal order.
Because surface geometry is unaffected by vehicle heading, the signed offset $s$ remains invariant.
Candidate correspondences are therefore evaluated under both relative headings $h \in \{0^\circ, 180^\circ\}$.

Second, physical occlusions and the limited sensor field of view often prevent capturing a complete contour around the passage circumference.
To ensure that unobserved regions do not corrupt matching, each descriptor tracks its validly observed cells, while considering both directions of observation.
Pairs with insufficient overlap are discarded to avoid false associations on noisy fragments.

Third, sensor viewpoint differences and measurement noise can perturb the estimated reference contour.
Because surface offsets are measured along outward normals to this contour, errors in the contour parameter introduce systematic distortions to the descriptor.
Relative scan misalignment is corrected before descriptor aggregation.
We normalize the descriptor field by fitting and subtracting these variations over observed cells:
\begin{equation}
f(z, p) = \alpha(p) + \gamma(z) + z\bigl(\beta_1 n_1(p) + \beta_2 n_2(p)\bigr),
\label{eq:nuisance}
\end{equation}
where $\alpha(p)$ and $\gamma(z)$ denote circumferential and axial biases, and $\beta_1, \beta_2$ are tilt coefficients for $n_1(p),n_2(p)$, respectively.
These parameters are estimated via least squares, allowing fine structural variations to be compared after removing the fitted distortions.

\begin{algorithm}[!t]
\caption{\method{}: Landmark-Based Trajectory Correction}
\label{alg:landmark_pgo}
\small
\algrenewcommand\algorithmicrequire{\textbf{Input:}}
\algrenewcommand\algorithmicensure{\textbf{Output:}}
\begin{algorithmic}[1]
\Require Local submaps $\{\mathcal P_k\}$ at frame $k$,
odometry trajectory $\mathcal X^{\mathrm{odom}}$
\Ensure Corrected trajectory $\mathcal{X}^*$
\State Initialize loop set $\mathcal{L} \gets \emptyset$
\State Estimate axes and reference contours from $\{\mathcal{P}_k\}$
\State Detect boundaries and construct normalized surface fields
\For{each query boundary $i$}
    \For{each eligible previous boundary $j$}
        \State Evaluate both relative headings and extract local patches
        \State Compute patch distances
        \State Find the unique mutual nearest-neighbor matches
        \State Count inliers $K_{ij}$ via geometric verification
        \State Compute $S(i,j) = 2K_{ij}/(N_i + N_j)$
        \Statex \hspace{\algorithmicindent}$N_i,N_j$: numbers of extracted keypoints
    \EndFor
    \If{a unique highest-scoring candidate $j^*$ exists}
        \If{$K_{ij^*} \ge K_{\mathrm{thresh}}$}
            \State Add $(i,j^*)$ to $\mathcal L$ using residual \Cref{eq:residual}
        \EndIf
    \EndIf
\EndFor
\State Obtain $\mathcal{X}^*$ by optimization in \Cref{sec:verification}
\Statex \hspace{\algorithmicindent}(keep $\mathcal{X}^*=\mathcal{X}^{\rm odom}$ if $\mathcal{L}=\emptyset$)
\State \Return Corrected trajectory $\mathcal{X}^*$
\end{algorithmic}
\end{algorithm}

After normalizing the field around keypoints, we apply a Gaussian kernel with standard deviations $\sigma_z$ and $\sigma_p$ to calculate the weighted difference of local patches.
The resulting mutual nearest-neighbor matches undergo geometric verification in the following section.

\subsection{Verification and Optimization}
\label{sec:verification}

Mutual nearest-neighbor patch matching provides tentative correspondences for candidate boundaries.
To filter out false positives before incorporating loop closures into the factor graph, a local geometric verification step confirms rigid spatial consensus among surface patches around the detected boundary.
The geometric verification is performed for each candidate pair,
yielding the consensus score $S(i,j)=2K_{ij}/(N_i+N_j)$,
where $K_{ij}$ is the number of geometric inliers and
$N_i,N_j$ are the numbers of extracted keypoints.
Candidates are ranked by consensus score, and the unique
highest-scoring candidate is accepted only if
$K_{ij} \ge K_{\mathrm{thresh}}$, where $K_{\mathrm{thresh}}$ is the minimum inlier count.

The gravity reference used for contour alignment can also
determine rotation about the corridor axis. However, errors
in this reference can introduce inaccurate rotational constraints.
We therefore use a 5-DoF geometric factor that constrains
cross-section positions and corridor axes while leaving rotation
about the degenerate axis unconstrained.
For each verified landmark pair $(i,j)$, let $k_i,k_j$ denote
the corresponding submap indices.
Let $\bar{\vect c}_i,\bar{\vect c}_j$ denote the vectors from
the respective submap origins to the observed contour centroids,
and $\vect a_i,\vect a_j$ the corresponding unit corridor axes.
These vectors are expressed in frame $k_i$ under the current
relative pose.
With $\vect t_{ij}$ denoting the translation of
$T_{k_i}^{-1}T_{k_j}$, the position and axis alignment errors are

\begin{equation}
	\begin{aligned}
		\vect r_{ij}
		&=\vect t_{ij}
		-\left(\bar{\vect c}_i-\bar{\vect c}_j\right),\\
		\delta_{ij}
		&=1-(\vect a_i^\top\vect a_j)^2.
		\label{eq:residual}
	\end{aligned}
\end{equation}

Given the initial odometry
$\mathcal X^{\mathrm{odom}}=\{T_k^{\mathrm{odom}}\}$,
we compute the relative odometry as
$\hat T^{\mathrm{odom}}_{k,k+1}
=\left(T_k^{\mathrm{odom}}\right)^{-1}
T_{k+1}^{\mathrm{odom}}$.
We then optimize the trajectory $\mathcal X=\{T_k\}$ by

\begin{equation}
	\begin{aligned}
		\mathcal X^{\ast}=\operatorname*{arg\,min}_{\mathcal X}\quad
		&\sum_k
		\!||
		T_k^{-1}T_{k+1} -
		\hat T^{\mathrm{odom}}_{k,k+1}
		||\\
		&+\sum_{(i,j)\in\mathcal L}
		\rho\!\left(\|\vect r_{ij}\|+|\delta_{ij}|\right)
		,
	\end{aligned}
	\label{eq:pgo_objective}
\end{equation}

where $\mathcal L$ contains the verified landmark pairs, and $\rho(\cdot)$ is a robust Huber loss.
The first term preserves relative motion derived from the
initial odometry trajectory.
The second term enforces agreement between the observed
cross-section positions and corresponding corridor axes,
leaving rotation about the common axis unconstrained.

\Cref{alg:landmark_pgo} summarizes the complete \method{} pipeline, from landmark detection and invariant patch matching to factor graph trajectory optimization.

\begin{figure*}[!t]
	\centering
	\begin{minipage}[t]{0.385\textwidth}
		\centering
		\includegraphics[width=\linewidth]{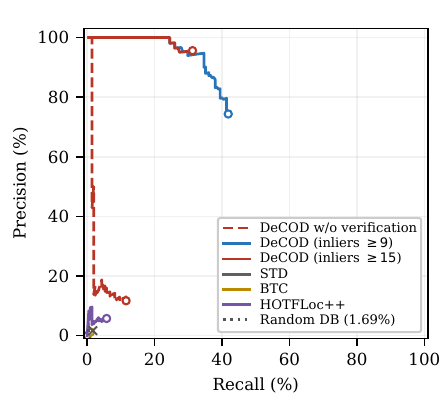}
		\par{\small (a)}
	\end{minipage}%
	\begin{minipage}[t]{0.60\textwidth}
		\centering
		\includegraphics[width=\linewidth]{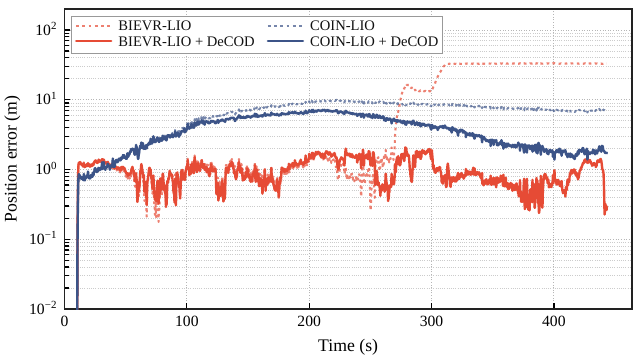}
		\par{\small (b)}
	\end{minipage}
	\vspace{1mm}
    \caption{Evaluation on GEODE. (a) Precision--recall curves on \textit{Shield} at 2.4\,m axial tolerance. (b) Position error on \textit{Shield S10-Beta} on a logarithmic scale: BIEVR-LIO (red) and COIN-LIO (blue), before (dotted) and after (solid) \method{} correction. To illustrate the pose drift over time, the trajectory origins were matched to ground truth at the start of the return traversal. Only rotation was fitted by least squares for visual comparison.}
	\label{fig:results}
\end{figure*}

\section{Experiments and Results}
\label{sec:protocol}\label{sec:evaluation}

We evaluated \method{} on the public GEODE benchmark~\cite{chen2024geode} and a custom underground pipe dataset across multiple LiDAR-inertial frontends.
We compared \method{} with STD~\cite{yuan2023std}, BTC~\cite{yuan2024btc}, and HOTFLoc++~\cite{griffiths2026hotfloc} on the same local point clouds.
Our experiments analyzed (i) landmark retrieval and axial localization under structural repetition, (ii) longitudinal drift correction across frontends, and (iii) component contributions alongside computational efficiency.

\subsection{Evaluation Setup}
\label{sec:experimental_setup}\label{sec:settings}\label{sec:reference}

Subterranean inspection typically operates without external positioning, making return traversals a primary opportunity for loop closure.
We built a landmark database during the outbound traversal and queried it during the return traversal.
Verified boundary associations introduce position and axis alignment constraints that connect the two traversals and reduce accumulated longitudinal drift through joint optimization with odometry factors.

\subsubsection{Datasets}
GEODE provides subterranean tunnel sequences spanning hundreds of meters, collected using Velodyne, Ouster, and Livox Avia LiDARs on devices \textit{Alpha}, \textit{Beta}, and \textit{Gamma}, respectively, together with synchronized IMUs.
We used ten \textit{Shield} sequences and seven \textit{Tunneling} recordings under diverse surface and moisture conditions.
The custom pipe dataset contains seven inspection sequences spanning tens of meters inside underground polyethylene and concrete pipes with diameters from 400 to 800\,mm, collected by a wheeled robot equipped with a Livox LiDAR and an IMU.

\subsubsection{Descriptor Extraction}
We constructed the point cloud $\mathcal P$ at 8\,s intervals by accumulating 20 consecutive LiDAR scans with motion compensation from LiDAR--inertial odometry and then voxelized it at 5.5\,cm resolution.
In this process, scan-wise correction was applied to reduce residual distortion across accumulated scans before descriptor construction.
The corridor axis $\vect a$ was then estimated from each local submap via principal component analysis (PCA).
Along this corridor axis, structural boundary candidates were detected using an axial DoG filter with standard deviations of 2.5\,cm and 10\,cm, followed by 0.8\,m nonmaximum suppression.
Before descriptor extraction, candidate landmarks were validated by requiring a ratio $\lambda_1/\lambda_2 \ge 2.0$ between the two largest
singular values to confirm an elongated corridor and accepting only contours where at least 90\% of the projected points lay within 0.18\,m of the reference contour $\vect c(p)$.
Around each validated landmark, the descriptor was formed on a $16\times256$ grid spanning $\pm2.4$\,m axially with 0.30\,m cell resolution and 256 normalized arclength bins along the contour.
In our setup, the descriptor was rotationally aligned with gravity from the IMU, and the bottom $\pm30^\circ$ sector was masked to suppress ground returns.

\begin{table*}[!b]
\centering
\begin{minipage}[t]{\dimexpr(\textwidth-\columnsep)/2\relax}
\vspace{0pt}
\centering
\caption{Place recognition performance on GEODE Shield 1--10. Recall@1 is evaluated over all common queries, \newline whereas Precision@1 is evaluated over accepted queries. \newline Recall@1 and Precision@1 are reported as percentages, and accepted queries as counts. The highest recall and precision in each column are in bold.}
\label{tab:retrieval}
\footnotesize
\setlength{\tabcolsep}{2.5pt}
\begin{tabular*}{\linewidth}{@{\extracolsep{\fill}}lccccc@{}}
\toprule
\multirow{2}{*}{Method} & \multicolumn{2}{c}{Recall@1} & \multicolumn{2}{c}{Precision@1} & \multicolumn{1}{c}{\multirow{2}{*}{\shortstack[c]{Accepted\\Queries}}} \\
\cmidrule(lr){2-3}\cmidrule(lr){4-5}
 & 0.3\,m & 2.4\,m & 0.3\,m & 2.4\,m & \\
\midrule
\method{} (w/o verification) & 5.8 & 11.5 & 5.9 & 11.7 & 205 \\
\method{} (inliers $\ge 9$) & \textbf{20.7} & \textbf{41.8} & 36.8 & 74.4 & 117 \\
\method{} (inliers $\ge 15$) & 17.3 & 31.3 & \textbf{52.9} & \textbf{95.6} & 68 \\
STD~\cite{yuan2023std} & 0.0 & 0.0 & 0.0 & 0.0 & 8 \\
BTC~\cite{yuan2024btc} & 0.0 & 1.0 & 0.0 & 1.2 & 170 \\
HOTFLoc++~\cite{griffiths2026hotfloc} & 1.9 & 5.8 & 1.9 & 5.8 & 208 \\
\bottomrule
\end{tabular*}
\end{minipage}\hfill
\begin{minipage}[t]{\dimexpr(\textwidth-\columnsep)/2\relax}
\vspace{0pt}
\centering
\caption{Place recognition performance on GEODE Tunneling 1--5 Gamma. Recall@1 is evaluated over all common queries, whereas Precision@1 is evaluated over accepted queries. Recall@1 and Precision@1 are reported as percentages, and accepted queries as counts. The highest recall and precision in each column are in bold.}
\label{tab:retrieval_tunneling}
\footnotesize
\setlength{\tabcolsep}{2.5pt}
\begin{tabular*}{\linewidth}{@{\extracolsep{\fill}}lccccc@{}}
\toprule
\multirow{2}{*}{Method} & \multicolumn{2}{c}{Recall@1} & \multicolumn{2}{c}{Precision@1} & \multicolumn{1}{c}{\multirow{2}{*}{\shortstack[c]{Accepted\\Queries}}} \\
\cmidrule(lr){2-3}\cmidrule(lr){4-5}
 & 0.3\,m & 2.4\,m & 0.3\,m & 2.4\,m & \\
\midrule
\method{} (w/o verification) & \textbf{39.1} & 52.2 & 39.1 & 52.2 & 23 \\
\method{} (inliers $\ge 9$) & \textbf{39.1} & \textbf{56.5} & 64.3 & \textbf{92.9} & 14 \\
\method{} (inliers $\ge 15$) & \textbf{39.1} & 47.8 & \textbf{75.0} & 91.7 & 12 \\
STD~\cite{yuan2023std} & 0.0 & 0.0 & 0.0 & 0.0 & 7 \\
BTC~\cite{yuan2024btc} & 4.3 & 21.7 & 8.3 & 41.7 & 12 \\
HOTFLoc++~\cite{griffiths2026hotfloc} & 0.0 & 0.0 & 0.0 & 0.0 & 23 \\
\bottomrule
\end{tabular*}
\end{minipage}
\end{table*}

\setcounter{topnumber}{1}
\begin{table}[t]
\centering
\caption{Trajectory position RMSE [m] on the GEODE benchmark with BIEVR-LIO base odometry.}
\label{tab:trajectory}
\footnotesize\setlength{\tabcolsep}{2.5pt}
\begin{tabular*}{\linewidth}{@{\extracolsep{\fill}}lrrrr@{}}
\toprule
Recording & No loop & BTC & HOTFLoc++ & \method{}\\
\midrule
S1-Gamma & 0.868 & 101.785 & $^{*}$103.340 & \textbf{0.308} \\
S2-Gamma & 0.493 & 89.405 & 94.418 & \textbf{0.474} \\
S3-Gamma & 1.276 & 70.046 & 75.233 & \textbf{1.121} \\
S4-Gamma & 2.497 & $^{*}$65.317 & $^{*}$73.851 & \textbf{2.142} \\
S5-Gamma & \textbf{2.061} & \textbf{2.061} & 33.817 & {\setlength{\fboxsep}{0pt}\colorbox{black!12}{\strut \textbf{2.061}}} \\
S6-Gamma & \textbf{0.519} & 31.251 & $^{*}$30.477 & 0.522 \\
S7-Beta & 45.227 & $^{*}$87.340 & 83.621 & \textbf{4.527} \\
S8-Beta & 61.597 & 104.057 & 101.951 & \textbf{0.768} \\
S9-Beta & 94.902 & 77.349 & $^{*}$76.803 & \textbf{24.819} \\
S10-Beta & 132.817 & $^{*}$94.812 & 95.377 & \textbf{4.206} \\
T5-Alpha & \textbf{0.164} & 2.894 & 21.749 & {\setlength{\fboxsep}{0pt}\colorbox{black!12}{\strut \textbf{0.164}}} \\
T5-Beta & \textbf{0.126} & 15.973 & 14.537 & {\setlength{\fboxsep}{0pt}\colorbox{black!12}{\strut \textbf{0.126}}} \\
T5-Gamma & 0.097 & 10.961 & 21.243 & \textbf{0.095} \\
\bottomrule
\end{tabular*}
{\raggedright\scriptsize $^*$Asterisks denote reaching the maximum optimization iterations. \\ Gray shading indicates outputs unchanged due to zero accepted factors.\\Bold denotes the lowest error.\par}
\end{table}

\begin{table}[b]
	\centering
	\caption{Endpoint error before and after correction on pipe sequences.}
	\label{tab:custom_pipe}
	\footnotesize\setlength{\tabcolsep}{4pt}
	\begin{tabular*}{\linewidth}{@{\extracolsep{\fill}}lrrr@{}}
		\toprule
		Sequence & Odometry [m] & Corrected [m] & Reduction [\%] \\
		\midrule
		\textit{hume\_01} & 13.342 & 0.013 & 99.9 \\
		\textit{hume\_02} & 0.189 & 0.028 & 85.4 \\
		\textit{hume\_03} & 1.738 & 1.217 & 30.0 \\
		\textit{hume\_04} & 0.544 & 0.022 & 95.9 \\
		\textit{pe\_01} & 3.103 & 0.720 & 76.8 \\
		\textit{pe\_02} & 15.730 & 0.409 & 97.4 \\
		\textit{pe\_03} & 3.320 & 0.187 & 94.4 \\
		\bottomrule
	\end{tabular*}
\end{table}

\subsubsection{Matching and Optimization}
For each query landmark, we matched $9\times33$ descriptor patches against database candidates.
Unique mutual nearest-neighbor patch correspondences underwent geometric verification in the frame aligned with gravity and the corridor axis, yielding geometric consensus scores $S(i,j)$ (\Cref{alg:landmark_pgo}).
For each query, we accepted the unique candidate with the highest
geometric consensus score only if its inlier count met the threshold; otherwise, the query was rejected.
We evaluated thresholds of $K \ge 9$ and $K \ge 15$.
For trajectory correction, verified geometric factors combining position and axis alignment errors (\Cref{eq:residual}) were integrated with relative odometry factors from BIEVR-LIO~\cite{pfreundschuh2026bievr} or COIN-LIO~\cite{pfreundschuh2024coin} under a Huber loss with a threshold of 1.345.

\subsubsection{Evaluation Protocol and Metrics}
All methods used identical local point clouds, partitioned into an outbound database and return queries.
Submaps lacking ground-truth support were removed before candidate selection.
Eligible query--database pairs were separated by at least 30\,s and shared no scans.
Registration baselines (STD~\cite{yuan2023std} and BTC~\cite{yuan2024btc}) processed inputs at 0.10\,m and 0.25\,m voxel resolutions, respectively, while HOTFLoc++~\cite{griffiths2026hotfloc} was evaluated zero-shot.

Metric axial localization was evaluated at tolerances of 0.3\,m and 2.4\,m, representing fine and coarse accuracy, respectively.
Because the ground truth provides continuous trajectories without loop correspondence labels, loop accuracy was assessed against these trajectories.
This assessment used the absolute boundary plane residual for \method{} and the magnitude of the 3D translation error projected onto the query corridor axis for registration baselines.
An accepted association is classified as a true positive if its axial error is within tolerance $\tau$, and as a false positive otherwise.
Recall@1 is the number of true positive Top-1 associations divided by the total number of common queries, whereas Precision@1 is the same number of true positives divided by the total number of accepted queries.
Finally, full trajectory estimation was evaluated via position root mean square error (RMSE) against continuous ground truth on GEODE (\Cref{tab:trajectory}) and endpoint error on the custom pipe dataset (\Cref{tab:custom_pipe}), while position error over time during the return traversal illustrates drift reduction (\Cref{fig:results}(b)).

\subsection{Place Recognition Performance in GEODE}
\label{sec:descriptor_evaluation}

\Cref{tab:retrieval,tab:retrieval_tunneling} and \Cref{fig:results}(a) report landmark retrieval and axial localization performance on \textit{Shield} and \textit{Tunneling}.
To assess the contribution of geometric verification, we evaluated \method{} with and without this stage.
Without geometric verification, the best candidate from patch similarity was selected, and with verification, candidates were reranked by geometric consensus,
and the top-ranked candidate was accepted only if its inlier count met the threshold.
Verification therefore affects both candidate selection and acceptance, allowing recall to improve by reranking.

\Cref{tab:retrieval,tab:retrieval_tunneling} report verified \method{} results for both $K \ge 9$ and $K \ge 15$.
The results presented below use $K \ge 15$ for geometric verification.
The \textit{Shield} sequences traverse tunnels assembled from repeated concrete rings, forming hundreds of meters of repetitive cylindrical passages.

In this setting, \method{} unrolls local surface geometry around detected boundaries into descriptors for local patch matching.
On \textit{Shield}, verification raised precision from 11.7\% to 95.6\% at 2.4\,m tolerance, whereas registration baselines achieved at most 5.8\% precision (\Cref{tab:retrieval}).

Unlike the smooth concrete rings of \textit{Shield}, the \textit{Tunneling} sequences contain irregular excavated rock walls and support structures (\Cref{tab:retrieval_tunneling}).
The forward-facing Livox LiDAR on \textit{Gamma} has a limited field of view.
Even under this partial coverage, verification raised precision from 52.2\% to 91.7\% at 2.4\,m while reducing accepted queries from 23 to 12, substantially reducing false associations.
Precision at 0.3\,m was 75.0\%.

High precision is important for SLAM pose-graph optimization, where false-positive constraints can corrupt the trajectory.

\subsection{Trajectory Correction Performance in GEODE}
\label{sec:drift_correction}

We evaluated trajectory correction on ten \textit{Shield} recordings and all three sensor recordings of \textit{Tunneling 5}.
BIEVR-LIO~\cite{pfreundschuh2026bievr} served as base odometry for all evaluated recordings, and COIN-LIO~\cite{pfreundschuh2024coin} was additionally evaluated on \textit{Shield S10-Beta} where BIEVR-LIO exhibited severe open-loop drift.
Relative transformations estimated by BTC~\cite{yuan2024btc} and HOTFLoc++~\cite{griffiths2026hotfloc} were incorporated as full relative pose factors, whereas \method{} contributed verified position and axis alignment constraints (\Cref{eq:residual}) with at least nine inliers ($K \ge 9$) in this trajectory evaluation.
STD~\cite{yuan2023std} was excluded from the trajectory comparison because of its near-zero axial localization performance in \Cref{tab:retrieval,tab:retrieval_tunneling}.
Graphs were optimized following \Cref{eq:pgo_objective} using Levenberg--Marquardt.

\method{} substantially reduced position RMSE on \textit{Shield S8-Beta} and \textit{S10-Beta} (\Cref{tab:trajectory}).
On \textit{S8-Beta}, position RMSE dropped from 61.597\,m to 0.768\,m.
On \textit{S10-Beta}, it dropped from 132.817\,m to 4.206\,m.
\Cref{fig:results}(b) further illustrates the reduction in position error on \textit{Shield S10-Beta} for BIEVR-LIO and COIN-LIO, showing that applying the verified geometric constraints reduced longitudinal drift across both frontends.

\method{} preserved the odometry accuracy across \textit{Tunneling~5} despite sensor variations across omnidirectional LiDARs (\textit{Alpha}, \textit{Beta}) and a LiDAR with a limited forward field of view (\textit{Gamma}), by effectively rejecting the false candidates. In cases where none of the proposed matches met the verification threshold (shaded gray in \Cref{tab:trajectory}), no loops were accepted.

This conservative acceptance policy, however, limits correction
when severe odometric drift coincides with insufficient reliable landmarks.
On \textit{Shield S9}, RMSE decreased from 94.902\,m to 24.819\,m,
leaving substantial trajectory error even after optimization due to the small number of loop closures.
This indicates that there still exists a margin for trade-off between rejecting unreliable
associations and retaining sufficient constraints for drift correction.

\subsection{Experiment on Custom Dataset}
\label{sec:application}

We evaluated \method{} on a wheeled inspection robot inside underground drainage conduits, examining precast concrete (\textit{hume}) and ribbed polyethylene (\textit{pe}) pipes.
Because pipe conduits with diameters from 400 to 800\,mm are an order of magnitude smaller than subterranean tunnels, we scaled the spatial point sampling size down by a factor of ten while maintaining the descriptor grid dimension at $16\times256$.
Along these uniform cylindrical conduits, our detector captured the annular joint seams between pipe segments.
In each trial, the robot performed an outbound and return traversal.
For evaluation, the endpoint error was computed as the Euclidean distance between the estimated trajectory endpoint and the measured endpoint, both expressed in the launch manhole coordinate frame.

\begin{figure}[!t]
    \centering
    \begin{minipage}[t]{0.5\linewidth}
        \centering
        \includegraphics[width=\linewidth]{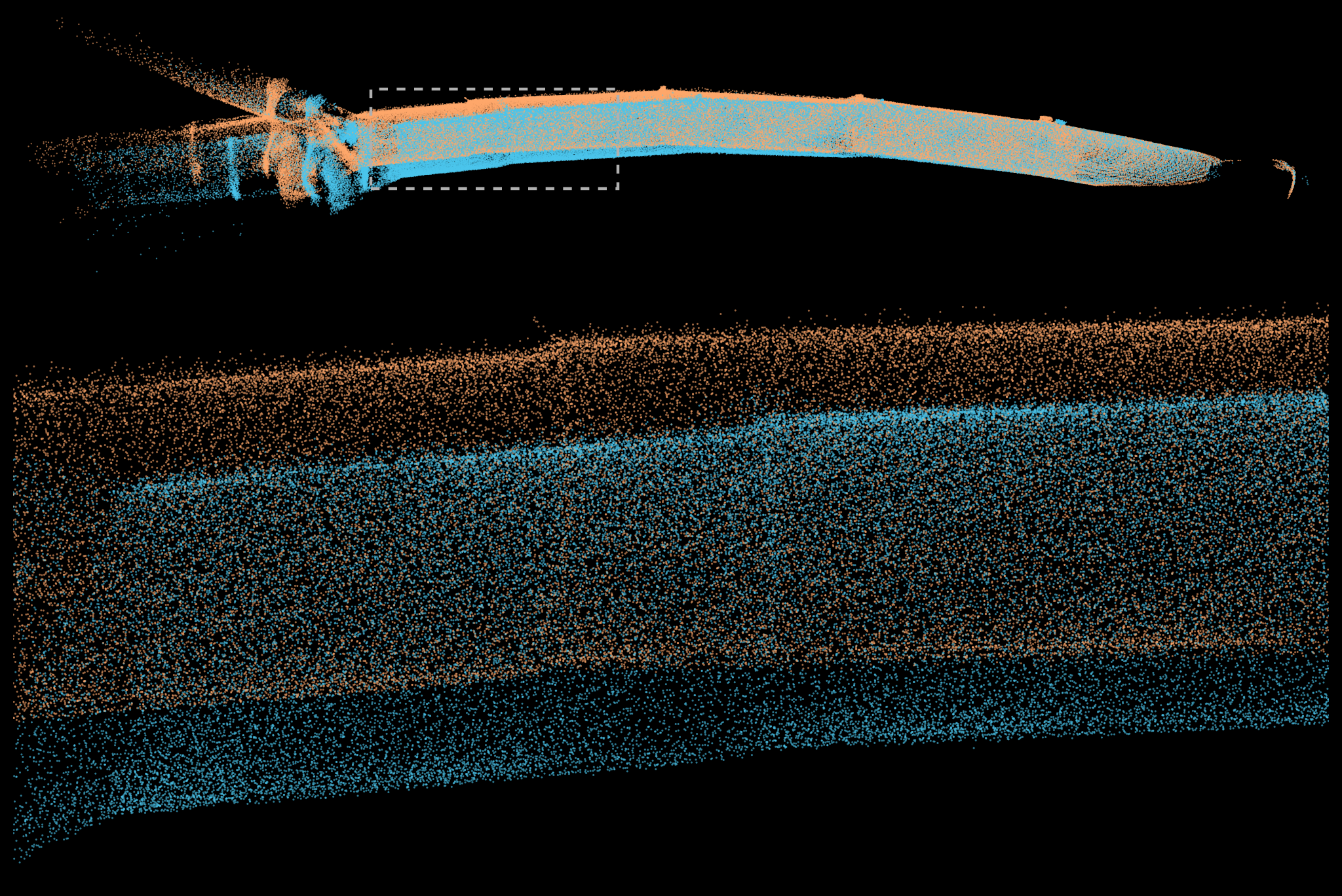}
        \par{\small (a)}
    \end{minipage}%
    \begin{minipage}[t]{0.5\linewidth}
        \centering
        \includegraphics[width=\linewidth]{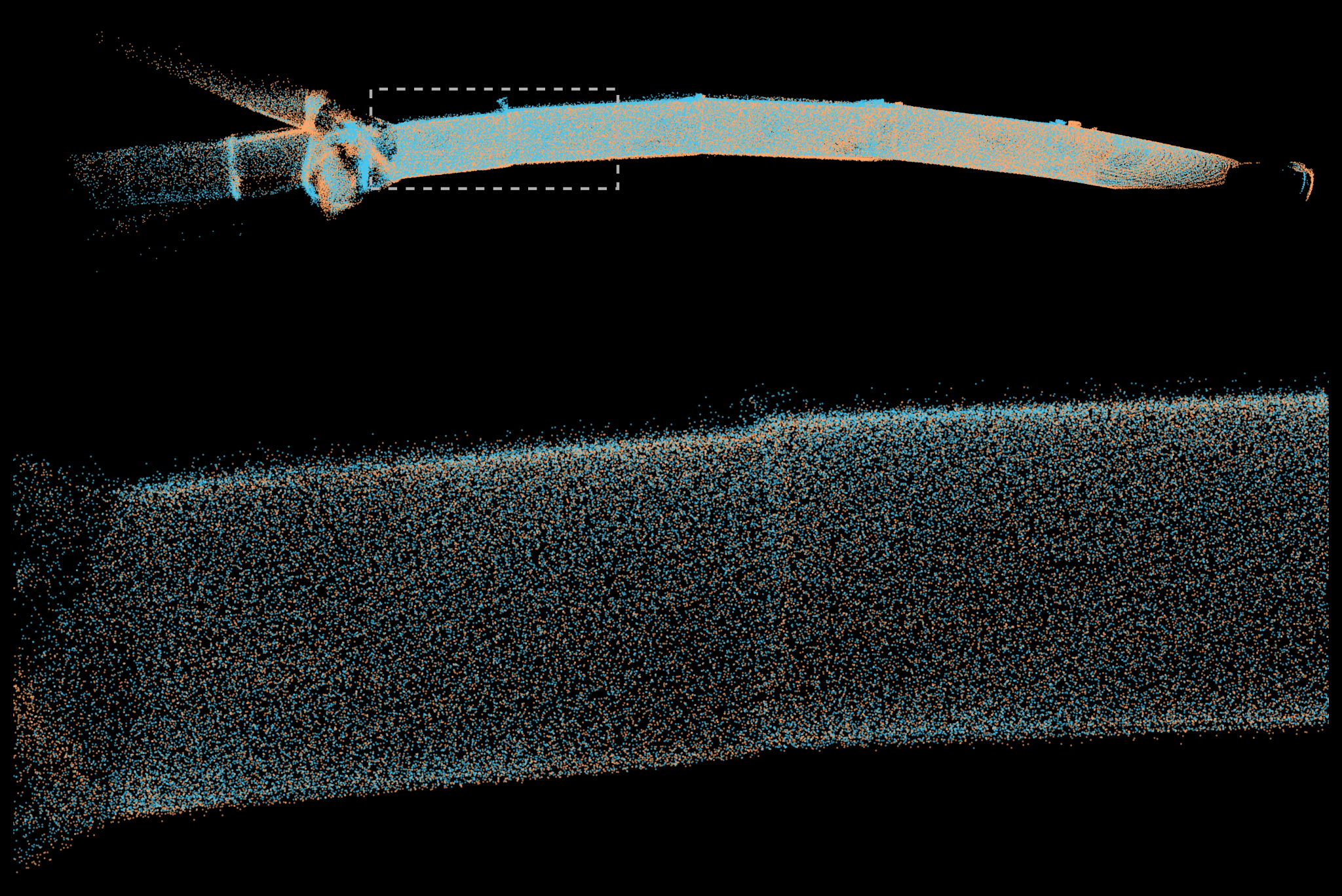}
        \par{\small (b)}
    \end{minipage}
    \caption{Point-cloud maps of \textit{hume\_04} reconstructed using (a) odometry and (b) odometry with \method{}. Cyan and orange show outbound and return scans. The boxed regions are magnified below, showing that odometry places the same pipe wall at two different positions due to pose drift, producing separate cyan and orange outlines. After our suggested correction based on loop closures, the outlines largely overlap and form a better reconstruction.}
    \label{fig:custom_pipe_maps}
\end{figure}

\Cref{tab:custom_pipe} reports endpoint errors across the pipe recordings.
Base odometry was BIEVR-LIO~\cite{pfreundschuh2026bievr} on \textit{hume} pipes and FAST-LIO2~\cite{xu2021fastlio2} on \textit{pe} pipes.
Incorporating verified geometric constraints at pipe joints substantially reduced endpoint errors across all sequences.
On \textit{hume} pipes, \method{} brought endpoint errors down to sub-meter levels in three of the four trials, reaching 0.022\,m on \textit{hume\_04}, with a reduction of up to 99.9\% on \textit{hume\_01}.
\Cref{fig:custom_pipe_maps} shows reduced wall thickness on \textit{hume\_04} as the outbound and return scans align more closely after correction.
On \textit{pe} pipes, consistent endpoint error reductions were achieved across all runs, reaching 94.4\% on \textit{pe\_03}.
Although LiDAR odometry suffered from severe drift without paramter tuning in these narrow \textit{pe} pipes,
the results show that modeling degeneracy, detecting features orthogonal to the degenerate direction, and incorporating them as constraints can still correct the fraction of this drift.

\subsection{Design and Performance Analysis}
\label{sec:ablation}\label{sec:runtime}

To understand how \method{} addresses structural degeneracy, we examined descriptor construction, geometric verification, and the choice of 5-DoF constraints, followed by runtime performance on GEODE.

We evaluated descriptor variants using Recall@1 on 80 \textit{Shield} queries with ground-truth matches at 0.75\,m axial tolerance.
We evaluated descriptor normalization separately from local patch normalization, using a five-variable model to compensate for contour centering and sensor tilt errors.
Replacing it with a scale-only correction reduced Recall@1 from 16.3\% to 12.5\%.
In a separate comparison, keeping only the sign of each offset
reduced Recall@1 to 10.0\%, showing that offset magnitudes also contribute to performance.

At the association stage, the \textit{Shield} results in \Cref{tab:retrieval} show the effect of geometric verification at 2.4\,m axial tolerance.
With $K \ge 9$, precision increased from 11.7\% to 74.4\% and Recall@1 from 11.5\% to 41.8\%, while accepted queries decreased from 205 to 117.
Verification thus improves candidate selection through geometric consensus among local patches while rejecting proposals with insufficient inliers.

For trajectory correction, we compared the proposed 5-DoF constraints with full 6-DoF relative pose (Pose3) factors, keeping matching, position constraints, weights, and evaluation settings unchanged.
Although Pose3 factors using odometry-derived rotations yielded similar RMSE on most sequences, RMSE increased from 0.768 to 6.403\,m on \textit{S8-Beta} and from 4.206 to 9.708\,m on \textit{S10-Beta}.
These results support the 5-DoF formulation, which avoids imposing a potentially errornous rotation estimate about the corridor axis.

Finally, we measured the runtime of descriptor extraction, matching, and verification on an NVIDIA RTX 4070 Ti SUPER GPU.
Across 987 candidate pairs, these stages took an average of 2.15\,ms per pair.
For a query against a database of 20 candidates, the latency was 69.7\,ms, compared with 321.6\,ms for HOTFLoc++.
This sub-100\,ms query latency supports periodic background evaluation on accumulated submaps.

\section{Conclusion}
\label{sec:conclusion}

We presented \method{}, a geometric approach for place recognition and drift correction in degenerate subterranean corridors.
By extracting cross-sectional boundaries along the travel path, \method{} introduces geometric factors that constrain cross-section position and corridor axis alignment to reduce accumulated longitudinal drift, while leaving rotation about the common axis unconstrained.
Across tunnel benchmarks and field pipe inspections, combining normal-offset descriptors with local patch matching and geometric verification reduces false associations and trajectory error.

Without external positioning references, subterranean SLAM relies on relative loop constraints only.
Re-identifying structural boundaries upon a revisit bounds accumulated drift, at least preserving segments where odometry remains locally accurate. As our current formulation considers a single dominant weak axis along uniform passages,
extending these directional constraints to degeneracy in multiple degrees of freedom remains for future work.

\clearpage
\end{document}